\documentclass[conference]{IEEEtran}\IEEEoverridecommandlockouts
\usepackage{etoolbox}
\usepackage{enumitem}
\usepackage[ruled,vlined,linesnumbered]{algorithm2e}
 \usepackage{amsmath,amssymb}
 \usepackage{subfigure}
 \usepackage{graphicx,graphics,color,psfrag}
 \usepackage{cite,balance}
 \usepackage{caption}
 \allowdisplaybreaks
 \usepackage{accents}
 \usepackage{amsthm}
 \usepackage{bm}
 \usepackage[english]{babel}
 \usepackage{multirow}
 \usepackage{enumerate}
 \usepackage{cases}
 \usepackage{stfloats}
 \usepackage{dsfont}
 \usepackage{color,soul}
 \usepackage{amsfonts}
 \usepackage{cite,graphicx,amsmath,amssymb}
 \usepackage{subfigure}
 \usepackage{fancyhdr}
 \usepackage{hhline}
 \usepackage{graphicx,graphics}
 \usepackage{array,color}
 \usepackage{amsmath}
\usepackage{float}
\usepackage{amssymb}
\usepackage{amsmath}
\usepackage{amsthm}
\usepackage{amsfonts}
\usepackage{graphicx}
\usepackage{makecell}

\usepackage{epstopdf}
\usepackage{cite}
\usepackage{amsmath,bm}
\usepackage{subfigure}
\usepackage{graphicx}
\usepackage{color}
\usepackage{graphicx}
\usepackage{calc}
\usepackage{caption}

\begin{document}

\title{Diffusion Model-Based Data Synthesis Aided Federated Semi-Supervised Learning}
\author{Zhongwei Wang, Tong Wu, Zhiyong Chen, Liang Qian, Yin Xu, Meixia Tao\\
Cooperative Medianet Innovation Center, Shanghai Jiao Tong University, Shanghai, China\\
		Email: \{wzw0424001x, wu\_tong, zhiyongchen, lqian, xuyin, mxtao\}@sjtu.edu.cn}

\maketitle
\begin{abstract}
Federated semi-supervised learning (FSSL) is primarily challenged by two factors: the scarcity of labeled data across clients and the non-independent and identically distribution (non-IID) nature of data among clients. In this paper, we propose a novel approach, diffusion model-based data synthesis aided FSSL (DDSA-FSSL), which utilizes a diffusion model (DM) to generate synthetic data, bridging the gap between heterogeneous local data distributions and the global data distribution. In DDSA-FSSL, clients address the challenge of the scarcity of labeled data by employing a federated learning-trained classifier to perform pseudo labeling for unlabeled data. The DM is then collaboratively trained using both labeled and precision-optimized pseudo-labeled data, enabling clients to generate synthetic samples for classes that are absent in their labeled datasets. This process allows clients to generate more comprehensive synthetic datasets aligned with the global distribution. Extensive experiments conducted on multiple datasets and varying non-IID distributions demonstrate the effectiveness of DDSA-FSSL, e.g., it improves accuracy from 38.46\% to 52.14\% on CIFAR-10 datasets with 10\% labeled data.
\end{abstract}

\section{Introduction}
Federated Learning (FL) allows multiple clients to collaboratively train machine learning models without the need to share raw data directly. It not only protects data privacy but also enhances the utilization of diverse and distributed data resources across different locations, thereby emerging as a pivotal technology for edge artificial intelligence applications \cite{yin}. The main challenge in FL is the substantial heterogeneity in data distribution across clients, commonly referred to as non-independent and identically distributed (non-IID) data \cite{li2020federated}. This disparity can cause divergence between the global and local models, known as client drift \cite{clientdrift}. Additionally, the scarcity of labeled data presents another obstacle, as obtaining large amounts of labeled data is often time-consuming and expensive in many domains, whereas unlabeled data is typically more readily available \cite{fedavg+fixmatch}.

To address these challenges, researchers have turned to Federated Semi-Supervised Learning (FSSL). FSSL combines the advantages of semi-supervised learning, which can leverage unlabeled data, with the collaborative framework of FL. SemiFed \cite{fedavg+fixmatch} incorporates consistency regularization and pseudo-labeling techniques, utilizing consensus among multiple client models to generate high-quality pseudo-labels, showing effectiveness in heterogeneous data distribution. FedMatch \cite{fedmatch} explores two scenarios: labels-at-client and labels-at-server, by introducing inter-client consistency loss and parameter decomposition, outperforming simple combinations of FL and semi-supervised learning in both cases. FedDure \cite{exnoniid} presents an FSSL framework with dual regulators to manage non-IID data across and within clients. 

Another approach is to use deep learning-based data augmentation techniques to aided FL. The synthetic data aided FL (SDA-FL) \cite{fedwithgan} framework shares synthetic data generated by locally pre-trained generative adversarial networks (GANs), utilizing an iterative pseudo labeling mechanism to improve consistency across local updates and enhance global aggregation performance in both supervised and semi-supervised cases. In \cite{ganfed}, a global generator is collaboratively trained within the FL framework to produce synthetic data. FedDISC\cite{yang2024exploring} introduces pre-trained diffusion models (DMs) \cite{wutong} into FL, utilizing prototypes and domain-specific representations to generate high-quality synthetic datasets.

FSSL and SDA-FL offer distinct approaches to addressing the challenges of heterogeneous data distribution and limited labeled data, providing valuable insights from two different perspectives. However, directly combining FSSL with SDA-FL poses new challenges. Existing SDA-FL methods often rely on pre-trained large generative models \cite{yang2024exploring}, which may not be suitable due to potential domain mismatches between the pre-trained models and the specific tasks. Additionally, training generative models capable of producing high-quality data \cite{ganfed} in FSSL is particularly difficult because of the scarcity of labeled data for model training and validation. 

Motivated by the observation, we propose a novel approach called Diffusion Model-based Data Synthesis Aided Federated Semi-Supervised Learning (DDSA-FSSL) to address the challenges in FL. Specifically, to overcome the challenge of data scarcity, a global classifier is employed for pseudo-labeling amounts of unlabeled data, and a precision-driven optimization process is applied to refine these pseudo-labeled samples, enhancing their quality and reliability. Instead of using pre-trained generative models, a global DM is collaboratively trained using both the labeled data and optimized pseudo-labeled data, thus avoiding domain mismatch issues. The DM enables clients to generate synthetic data for absent classes in their local datasets, effectively addressing the challenge of heterogeneous data distribution. Experimental results show that DDSA-FSSL significantly enhances classification accuracy compared to existing methods. For example, with 10\% labeled CIFAR-10 data under dual heterogeneity, it raises accuracy from 38.46\% to 47.72\% using 10\% synthetic data, and to 53.01\% with 90\% synthetic data.
\begin{figure*}[t]
    \centering
    \includegraphics[width=\textwidth]{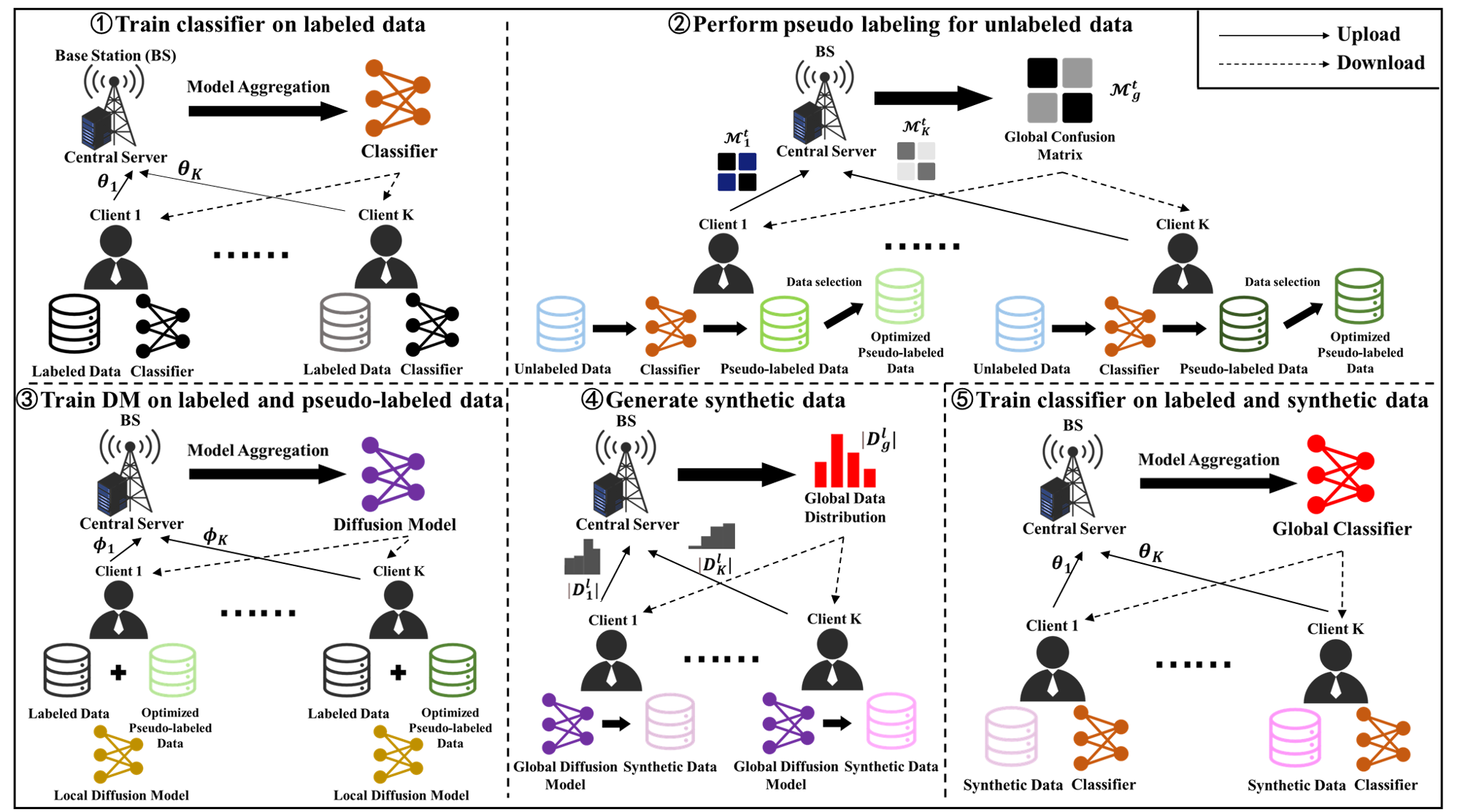}
    \caption{Overview of the proposed DDSA-FSSL. In the first step, each client performs federated training of a global classifier using labeled data. In the second step, the global classifier performs pseudo-labeling for the unlabeled data at each client, followed by a precision-driven optimization process guided by the global confusion matrix $\mathcal{M}_g^t$ to refine and select high-quality pseudo-labeled samples. In the third step, clients collaboratively train the DMs using both the labeled and optimized pseudo-labeled data. In the fourth step, the DMs are employed by clients to generate specific synthetic data, based on discrepancies between local and global data distributions. Finally, clients conduct federated training of the classifier using both labeled and synthetic data.
    } 
    \label{fig:system model}
\end{figure*}
\section{System Model}
We consider a FSSL scenario with heterogeneous data distributions, specifically focusing on the labels-at-clients setting \cite{fedmatch}. As shown in Fig. \ref{fig:system model}, the system consists of $K$ clients, each with their own local data, and one base station (BS) with a central server that coordinates the FL process without having direct access to the client's data. FL seeks to train a global classifier by coordinating these clients, each of which trains a local model using its own data. The central server then applies a federated aggregation algorithm to obtain the global parameters $\theta_g$.

For FSSL, each client's dataset consists of a labeled subset $D_k^{l}=\{ x_{k,i}^l, y_{k,i}\}_{i=1}^{|D_k^l|} = \bigcup_{c=1}^C (D_{k,c}^{l})$ and an unlabeled subset $D_k^u=\{ x_{k,i}^u\}_{i=1}^{|D_k^u|} = \bigcup_{c=1}^C (D_{k,c}^{u})$, where $x_{k,i}^l$ represents a labeled sample from client $k$, $y_{k,i}$ is its corresponding label, and $C$ denotes the total number of classes. In real-world applications, the assumption that all data is fully annotated is unrealistic due to the significant effort and cost associated with data labeling \cite{fedmatch}. Consequently, the ratio of labeled data, denoted as $\lambda = |D_k^l|/\left(|D_k^l| + |D_k^u| \right)$, is typically small. In heterogeneous data distribution scenarios, both the labeled and unlabeled datasets on each client may only contain samples from a subset of classes, with few or no samples from other classes. Moreover, the distributions of labeled and unlabeled data across different clients are also heterogeneous\cite{exnoniid}. This scarcity of labels and the heterogeneity of data distributions \cite{federatednoniid, exnoniid} can result in significant performance degradation in FL.

To address this challenge, we consider that each client is equipped with a local conditional latent diffusion model (c-LDM). Unlike existing methods that rely on unsupervised GANs \cite{fedwithgan}, which cannot generate samples for specific classes, c-LDM not only allows precise control over the synthesis of data for targeted classes, but also provides more stable training and generates higher-quality, more diverse synthetic data. While some approaches utilize pre-trained DMs like Stable Diffusion \cite{fedwithsdm}, these approaches have limitations in FSSL scenarios. Pre-trained models, trained on large-scale general datasets, often struggle to adapt to the client-specific data distributions and impose high computational demands, making them impractical for resource-constrained devices in FL environments.

\section{Diffusion Model-Based Data Synthesis Aided FSSL}\label{sec:method}
In this section, we introduce DDSA-FSSL framework that leverages DMs to generate synthetic data for classes. As show in Fig. \ref{fig:system model}, the proposed DDSA-FSSL is divided into five steps, which are detailed in the following subsections.

\subsection{Enhancing Dataset with Pseudo-Labels}
At the start of DDSA-FSSL, due to the scarcity of labeled data and the inherent complexity of high-dimensional parameter spaces in c-LDM, it is challenging to directly train a c-LDM capable of producing high-quality and diverse synthetic data. To leverage the unlabeled data $\{D_k^u\}^K_{k=1}$, clients first collaboratively train a global classifier using the FedAvg algorithm \cite{fedavg} based on the labeled data $\{D_k^l\}^K_{k=1}$. 

In FedAvg, the local objective function for client $k$ at each communication round $ r = 1,..., R$ is defined as follows:
\begin{align}
F_{k}\left(\boldsymbol{\theta}_{r,e}^{k}\right) \triangleq \mathbb{E}_{(x, y) \sim \mathcal{D}_{k}} \mathcal{L}_{c}\left(\boldsymbol{\theta}_{r,e}^{k} ; x, y\right),
\end{align}
where $\mathcal{L}_{c}$ is the cross-entropy and $\boldsymbol{\theta}_{r,e}^{k}$ denotes the parameters of client $k$ after $e$ local updates in communication round $r$. 

During each local training epoch $ e = 0,1,..., E$, each client updates its local parameters:
\begin{equation}
\boldsymbol{\theta}_{r, e+1}^k \leftarrow \boldsymbol{\theta}_{r, e}^k-\eta_r \nabla F_k\left(\boldsymbol{\theta}_{r,e}^{k}\right),\quad \boldsymbol{\theta}_{r}^k \leftarrow \boldsymbol{\theta}_{r, E}^k,
\end{equation}
where $\eta_r$ is the learning rate of round $r$.

The updated local parameters $\boldsymbol{\theta}_r^k$ for the $k$-th client are then sent back to the central server for parameters aggregation. The FedAvg algorithm  aims to aggregate the local parameters of clients based on the amount of training data each client contributes:
\begin{equation}
     \boldsymbol{\theta}_{r}^{g} \triangleq \sum_{k=1}^{K} p_k \boldsymbol{\theta}_{r}^{k}, \quad p_k= \frac{\left|{D}_{k}\right|}{\sum_{k=1}^{K}\left|{D}_{k}\right|},
\end{equation}
where $p_k$ is the aggregation weight for client $k$. 

When $e = 0$, $ \boldsymbol{\theta}_{r,0}^{k} = \boldsymbol{\theta}_{r-1}^{g}$, which indicates the beginning of each local training round where clients download the global parameters from the central server. 

Through the iterative process of local training and global aggregation, the central server is expected to converge to a global model with the global parameters $\theta_g$: 
\begin{equation}
    \theta_g = FedAvg(K, R, E, \{D_k^l\}_{k=1}^K, \mathcal{L}_{c}).
\end{equation}

The global classifier is used to perform pseudo labeling on the unlabeled data. The resulting pseudo-labeled datasets are denoted as $\{D_k^p\}_{k=1}^K = \{\{ x_{k,i}^u, \hat{y}_{k,i}\}_{i=1}^{|D_k^u|}\}_{k=1}^K$, where $\hat{y}_{k,i}$ denotes the corresponding pseudo-label of the unlabeled sample $x_{k,i}^u$. The data utilized for training the c-LDM comprises two components: labeled data $\{D_k^l\}_{k=1}^K$ and pseudo-labeled data $\{D_k^p\}_{k=1}^K$. Ideally, the c-LDM should accurately learn the distribution of the training data, enabling it to generate synthetic data that similar to the true distribution. However, the presence of mislabeled samples within the pseudo-labeled data leads to the generation of synthetic data that deviates from the true distribution, thus compromising the quality and reliability of the synthetic data. To mitigate this issue, we propose a data selection method based on precision optimization, which aims to filter pseudo-labeled data and enhance the quality and reliability of the data participating in the c-LDM training process.
\subsection{Precision-Optimized Data Selection}
The confusion matrix reflects the characteristics of the samples and their corresponding labels, therefore, we use the confusion matrix to measure the accuracy of pseudo-labels for samples predicted as specific classes. For $D_k^l$, the confusion matrix $\mathcal{M}_k^l$ is inherently a diagonal matrix, indicating perfect alignment between true and predicted labels. However, for $D_k^p$, the confusion matrix $\mathcal{M}_k^p$ needs to be estimated, as the true labels are not known with certainty. Specifically, each client generates the confusion matrix $\mathcal{M}_k^t$ by applying the global classifier to their local test set. These matrices are then uploaded to the central server, which aggregates them to construct a global confusion matrix $\mathcal{M}_g^t$ = $\sum_{k=1}^K \mathcal{M}_k^t$. Each client subsequently downloads $\mathcal{M}_g^t$ to estimate the $\mathcal{M}_k^p$ based on the principle that each column in $\mathcal{M}_g^t$ represents the distribution of true classes among samples predicted as a particular class. For each class $j$:
\begin{equation}
\mathcal{M}_k^p[:,j] = \frac{\mathcal{M}_g^t[:,j]}{\sum_{i=1}^{C} \mathcal{M}_g^t[i,j]} \cdot n_{k,j},
\end{equation}
where $n_{k,j}$ is the number of samples pseudo-labeled as class $j$ in the $D_k^p$, and $\mathcal{M}_k^p[:,j]$ denotes the $j$-th column of $\mathcal{M}_k^p$. 

Let $\boldsymbol{\rho}_{k}= (\rho_{k,1}, \ldots, \rho_{k,C})$ represents the proportion of each class in the $D_k^p$ selected by client $k$, the confusion matrix of the training data can be written as $\mathcal{M}(\boldsymbol{\rho}_{k}) = \mathcal{M}_{k}^l + \mathcal{M}_{k}^p \cdot diag(\boldsymbol{\rho}_k)$. Each client seeks to find the optimal selection of data that maximizes the average label precision: 
\begin{equation}
    \bar{P}_{k}(\boldsymbol{\rho}_{k}) = \frac{1}{|\mathcal{J}_{k}|} \sum_{j \in \mathcal{J}_{k}} \frac{\mathcal{M}(\boldsymbol{\rho}_{k})[j,j]}{\sum_{i=1}^C \mathcal{M}(\boldsymbol{\rho}_{k})[i,j]}, 
\end{equation}
where $\mathcal{J}_{k} = \{j : \sum_{i=1}^C \mathcal{M}(\boldsymbol{\rho}_{k})[i,j] \neq 0\}$. 

The optimization problem $\mathcal{P}_1$ can be formulated as:
\begin{align}
\max \limits_{\boldsymbol{\rho}_{k}} \quad &\bar{P}_{k}(\boldsymbol{\rho}_{k}) - w_{L_1} \sum_{c=1}^C |\rho_{k,c}| - w_{p}(\frac{1}{C} \sum_{c=1}^C \rho_{k,c} - \tau)^2 \label{Problem}\\ 
&s.t. \quad 0 \leq \rho_{k,c} \leq 1, \quad c = 1, \ldots, C, \tag{\ref{Problem}{a}} \label{Problema}
\end{align}
where $w_{L_1}$ denotes the weight for the $L_1$ regularization term,  $w_{p}$ denotes the weight for the penalty term and $\tau$ denotes the target proportion used to control the average selection proportion across all classes. 

The $L_1$ regularization promotes sparsity in the optimal solution and the penalty function facilitates a trade-off between the quantity of data employed and the precision of labels. $\mathcal{P}_1$ can be solved using sequential least squares programming (SLSQP). When the local optima proportion $\boldsymbol{\rho}_k^{\star}$ is obtained, each client randomly removes a corresponding proportion of data from each class in $D_k^p$, resulting in the optimized pseudo-labeled datasets $\{\hat{D}_k^p\}_{k=1}^{K}$.

\begin{algorithm}[t]
\small
\SetAlgoLined
\SetKwInOut{Input}{Input}
\SetKwInOut{Output}{Output}
\caption{\small Generate Class-conditional Synthetic Data}
\label{alg:generate synthetic data}

\Input{Local labeled data distribution $|D_{k}^l|$, global data distribution $|D_g^l|$, augmentation strength $\alpha$, timestep $T$ and global VAE's decoder $De(\cdot)$}
\Output{Synthetic dataset $D_{k}^{syn}$}


// Calculate number of synthetic samples to generate

\For{$c = 1$ \KwTo $C$}{


$|D_{k,c}^{syn}| = \max(0, \alpha(|D_{k}^l|+|D_{k}^{u}|) \cdot \frac{|D_{g,c}^l|}{|D_g^l|}-|D_{k,c}^{l}|)$
}

$D_{k}^{syn} = \{\}$

\For{$c = 1$ \KwTo $C$}{
\For{$i = 1$ \KwTo $|D_{k,c}^{syn}|$}{
$z_T \sim \mathcal{N}(0, I)$

//Denote $\alpha_t = 1-\beta_t$ and $\bar{\alpha}_t = \prod_{i=1}^t \alpha_i$

    \For{$t = T$ \KwTo $1$}{
        $\epsilon \sim \mathcal{N}(0, I)$ if $t > 1$, else $\epsilon = 0$\\
         \footnotesize$z_{t-1} = \frac{1}{\sqrt{\alpha_t}}\left(z_t-\frac{1-\alpha_t}{\sqrt{1-\bar{\alpha}_t}}\epsilon_{\phi_g}(z_t, t, c) \right)+ \sqrt{1-\bar{\alpha}_t} \epsilon$
    }
    
    $x^{syn} = De(z_0)$
    
    $D_{k}^{syn} = D_{k}^{syn} \cup \{(x^{syn}, c)\}$
}
}

\Return $D_{k}^{syn}$
\end{algorithm}

\subsection{Federated Diffusion}
The c-LDM used in DDSA-FSSL consists of three components: an encoder $En(\cdot)$, a decoder $De(\cdot)$, and a conditional diffusion model (CDM) $\mathcal{D}(\cdot)$ operating in the latent space. The encoder and decoder are implemented using a Variational Autoencoder (VAE) \cite{VAE}. The encoder encodes the input data $x$ into a latent representation $z = En(x)$, while the decoder reconstructs the data from the latent space, $\Tilde{x}={De}(z)={De}\left({En}(x)\right)$.

The CDM follows a two-phase process: a forward process and a reverse process. The forward process consists of a series of $T$ timesteps, during which Gaussian noise is gradually added to a clean latent representation $z_0$ according to a variance schedule $\bar{\beta}_{1:T}$. The reverse process is then trained to reconstruct the original $z_0$ by progressively removing the noise from $z_t$.
$z_t$ can be sampled in a single step given the $z_0$ and fixed variances:
\begin{equation}
z_t=\sqrt{1-{\bar{\beta_t}}}z_0+\sqrt{{\bar{\beta_t}}}\epsilon_t,\quad\epsilon_t\sim\mathcal{N}(0,I),
\end{equation}
where $ t \in [1,T]$ and $0<\bar{\beta}_{1:T}<1$. The reverse process involves training a neural network, typically U-Net, denoted as $\epsilon_{\phi}$ to serve as the noise predictor by estimating the noise $\epsilon_{t}$ at each timestep $t$. To enable each client to generate images of specific classes, we adopt a solution \cite{LDM2022} that incorporates a cross-attention mechanism into the intermediate layers of the U-Net network. The loss function can be written as:
\begin{equation}
     \mathcal{L}_{\textit{CDM}}=\mathbb{E}_{z_t,y,\epsilon\thicksim\mathcal{N}(0,1),t}\left[\|\epsilon_t-\epsilon_\phi\left(z_t,t,y\right)\|_2^2\right]\label{lcdm},
\end{equation}
where $y$ represents the label associated with $z_t$. 
Here, all components of the c-LDM are collaboratively trained through FL. Local training of the CDM alone would typically result in models that are unable to generate samples for missing classes. While \cite{fedwithgan} attempted to address this issue by uploading locally generated synthetic data to form a global synthetic dataset, their method requires substantial communication resources, as it involves transmitting entire synthesis datasets rather than just model parameters. Furthermore, the proposed method ensures a consistent latent space representation across all clients and allows for a better capture of the global data distribution. Consequently, all clients can achieve more accurate and diverse reconstructions, effectively transcending the limitations of their heterogeneous local data. 

The training of c-LDM comprises two stages. In the first stage, each client trains the VAE on both labeled and unlabeled data. The global parameters of the VAE $\Phi_g$ are obtained through FedAvg:
\begin{equation}
    \Phi_g = FedAvg(K, R, E, \{D_k^l \cup D_k^u\}_{k=1}^K  , \mathcal{L}_{\textit{VAE}}),
\end{equation}
where $\cup$ denotes the union of sets. The training loss function for VAE includes several components to ensure high-quality reconstructions and realistic generations:
\begin{equation}
    \mathcal{L}_{\textit{VAE}}=\|\Tilde{x}-x \|+\mathcal{L}_{\textit{KL}}+\mathcal{L}_{\textit{perceptual}}+\mathcal{L}_{\textit{GAN}},
\end{equation}
where $\|\tilde{x}-x\|$ denotes the reconstruction loss, and $\mathcal{L}_{\textit{KL}}$ is the Kullback-Leibler divergence loss. The term $\mathcal{L}_{\textit{perceptual}}$ \cite{perceptualloss} captures high-level features and semantic information. $\mathcal{L}_{\textit{GAN}}$ \cite{vaegan} involves a discriminator network that attempts to distinguish between real and reconstructed images. 

In the second stage, each client trains the CDM in latent space. Specifically, each client uses the global encoder $En(\cdot)$ with $\Phi_g$ to encode both the labeled datasets $D_k^l$ and the optimized pseudo-labeled datasets $D_k^p$ into the latent representation space: $\{D_k^{en}\}_{k=1}^{K} = \{{En}(D_k^l \cup \hat{D}_k^p)\}_{k=1}^{K}$. This encoding stage preserves essential features of the original data while reducing the communication cost during FL of the CDM with global parameter:
\begin{equation}
    \phi_g = FedAvg(K, R, E, \{D_k^{en}\}_{k=1}^{K}, \mathcal{L}_{\textit{CDM}}).
\end{equation}

\subsection{Synthetic Data Augmentation}
To generate specific synthetic data that aligns local data distribution with the global data distribution, each client $k$ uploads its local data distribution $|D_{k}^l|=\sum_{c=1}^{C}|D_{k,c}^{l}|$ to the central server. The server aggregates these distributions to construct the global data distribution $|D_g^l|=\sum_{k=1}^{K}\sum_{c=1}^{C}|D_{k,c}^{l}|$, which is then downloaded by each client. 
To measure the degree of synthetic data augmentation, we introduce a variable, denoted as augmentation strength $\alpha$:
\begin{equation}
\alpha = \frac{|D_{k}^l|+|D_{k}^{syn}|}{|D_{k}^l|+|D_{k}^{u}|}\label{alpha}, \\
\end{equation}
where $D_{k}^{syn}$ denotes the synthetic data generated by client $k$. 

The augmentation strength $\alpha$ determines the quantity of data in the synthetic dataset $D_{k}^{syn}$. Given the constraint that the distribution of the augmented local dataset ($D_{k}^{syn} \cup D_{k}^l$) matches the global data distribution $D_g^l$, $D_{k}^{syn}$ can be determined by (\ref{alpha}). It is important to note that in cases of extreme data imbalance, such as when a client's dataset contains only one or two classes and lacks data from the remaining classes, (\ref{alpha}) and ($D_{k}^{syn} \cup D_{k}^l) \sim D_g^l \label{syn}$ cannot be simultaneously satisfied. To address these scenarios, we implement a two-phase strategy. First, each client computes the total size of the combined synthetic and labeled datasets using (\ref{alpha}). Next, the client determines the amount of data for each class that aligns with the global distribution $D_g^l$. We denote this target distribution as $D_{k,c}^{l+syn}$, which ensures compliance with the following constraints:
\begin{equation}
\begin{aligned}
\bigcup_{c=1}^C (D_{k,c}^{l+syn})\sim D_g^l \ \text{and} \
\sum_{c=1}^C (|D_{k,c}^{l+syn}|) =|D_{k}^l|+|D_{k}^{syn}|.
\end{aligned}
\end{equation}

\addtolength{\topmargin}{0.05in}
\begin{table*}[t]
\centering
\caption{Performance comparison of DDSA-FSSL on two different data heterogeneity settings and different augmentation strength $\alpha$.}
\label{table:performance_comparison}
\renewcommand{\arraystretch}{1.2}
\resizebox{\textwidth}{!}{%
\begin{tabular}{|c|c|c|c|c|c|c|c|}
\hline
\multicolumn{2}{|c|}{\multirow{2}{*}{\small Methods}} & \multicolumn{3}{c|}{\small CIFAR10 ($\lambda = 0.1$)} & \multicolumn{3}{c|}{\small Fashion-MNIST ($\lambda = 0.1$)} \\
\cline{3-8}
 \multicolumn{2}{|c|}{} & \small (IID, IID) & \small (IID, DIR) & \small (DIR, DIR) & \small (IID, IID) & \small (IID, DIR) & \small (DIR, DIR) \\
\hline
\multicolumn{2}{|c|}{\small FedAvg} & \small 46.96\% & \small 46.85\% & \small 38.46\% & \small 87.21\% & \small 86.75\% & \small 70.01\% \\
\multicolumn{2}{|c|}{\small FedAvg-SL} & \small 73.72\% & \small 73.60\% & \small 63.02\% & \small 91.86\% & \small 91.62\% & \small 89.29\% \\
\hline
\multirow{5}{*}{\makecell[c]{\small DDSA-FSSL \\ (without/with \\data selection)}} &$\alpha = 0.2$& \small 53.62\%/54.94\% & \small 49.17\%/49.85\% & \small 44.31\%/47.72\% & \small 87.60\%/87.69\% & \small 86.91\%/86.98\% & \small 84.31\%/84.43\% \\
 &$\alpha = 1.0$& \small 56.48\%/60.22\% & \small 55.89\%/57.35\% & \small 48.74\%/53.01\% & \small 87.84\%/88.13\% & \small 87.23\%/87.30\% & \small 84.45\%/85.58\% \\
 &$\alpha = 2.1$& \small 58.51\%/62.58\% & \small 58.01\%/62.16\% & \small 49.15\%/53.98\% & \small 88.30\%/88.34\% & \small 87.29\%/87.44\% & \small 84.66\%/85.62\% \\
 &$\alpha = 4.1$& \small 59.86\%/63.34\% & \small 59.27\%/63.01\% & \small 50.99\%/55.22\% & \small 88.37\%/88.69\% & \small 87.64\%/87.97\% & \small 85.27\%/85.97\% \\
 &$\alpha = 10.1$& \small 61.15\%/\textbf{64.43\%} & \small 60.21\%/\textbf{63.72\%} & \small 51.37\%/\textbf{57.48\%} & \small 88.60\%/\textbf{89.03\%} & \small 88.14\%/\textbf{88.80\%} & \small 85.62\%/\textbf{86.24\%}\\
\hline
\end{tabular}%
}
\end{table*}
The amount of synthetic data to be generated for each classes can be calculated as:
\begin{equation}
    |D_{k,c}^{syn}| = \max\left(0, |D_{k,c}^{l+syn}|-|D_{k,c}^{l}|\right),
\end{equation}
which ensures that synthetic data is generated only when the desired amount $|D_{k,c}^{l+syn}|$ exceeds the available labeled data $|D_{k,c}^{l}|$ for a given class. The detail of generating synthetic data is outlined in Algorithm \ref{alg:generate synthetic data}.

\section{Experimental Results}
In this section, we present experimental results to verify the performance gain of the proposed DDSA-FSSL.
\subsection{Experimental Settings}
We conduct extensive experimental analyses on two distinct datasets: CIFAR-10 and Fashion-MNIST. To simulate the complex data distribution imbalances commonly encountered in real-world scenarios, the simulation incorporates two types of non-IID imbalances\cite{exnoniid}. 1) \textbf{External imbalance}: the labeled data distributions across different clients are heterogeneous. 2) \textbf{Internal imbalance}: within each client, the labeled and unlabeled data typically exhibit distinct distributions. The experiments can be divided into three scenarios based on the distribution of labeled and unlabeled data $(\{D_k^l\}_{k=1}^K, \{D_k^u\}_{k=1}^K)$: $(\textit{IID, IID})$, $(\textit{IID, DIR})$, and $(\textit{DIR, DIR})$. Here, $\textit{DIR}$ represents a non-IID scenario where data allocation follows a Dirichlet distribution $Dir(\gamma)$. We set the concentration parameter $\gamma$ to 0.1 across all datasets, determining the degree of data heterogeneity among the $K$ clients.

For the training of classifier, we use the ResNet-18 architecture as the default backbone. The Stochastic Gradient Descent (SGD) optimizer is employed with a momentum of 0.9, weight decay of $10^{-4}$, and a learning rate of $\eta_c$ = $10^{-4}$. For the VAE, we adopt the architecture outlined in \cite{vaegan}, with the encoder downsampling factor set to $f= H/h = W/w = 2$ \cite{LDM2022}. The VAE is optimized using the Adam optimizer with a learning rate of $\eta_v = 10^{-4}$. The exponential decay rates for the first and second moment estimates are set to 0.5 and 0.9, respectively. For the CDM, we employ a U-net convolutional neural network to approximate the predicted noise $\epsilon_{\phi}$. We use the diffusion parameters from \cite{DDPM2020} with $T = 1000$ timesteps and a linear noise schedule with $\beta_1 = 10^{-4}$ and $\beta_{1000} = 2\times10^{-2}$. Additionally, the Adam with weight decay (AdamW) optimizer is used with a learning rate of $\eta_d$ = $ 2\times10^{-4}$. In all FL scenarios, we adopt the FedAvg \cite{fedavg} algorithm, as it serves as a general approach. This choice is motivated by our focus on evaluating the impact of generated synthetic data on the system. It is worth noting that the parameter aggregation algorithm can be replaced with alternative algorithms tailored for FSSL.

\subsection {Results Analysis}
Table \ref{table:performance_comparison} presents a performance comparison of the proposed DDSA-FSSL against FedAvg and FedAvg-SL under different data heterogeneity settings and augmentation strengths. FedAvg-SL represents fully supervised training using FedAvg, where the entire dataset is labeled, serving as the upper bound for performance. In constrast, DDSA-FSSL and FedAvg are trained only on labled data with $\lambda=0.1$. The proposed DDSA-FSSL demonstrates performance improvements across all heterogeneous settings on both the  CIFAR-10 and Fashion-MNIST datasets. Meanwhile, we can observe that as the augmentation strength $\alpha$ increases, the classification accuracy progressively approaches the performance of FedAvg-SL. Furthermore, ablation studies indicate that our proposed precision-optimized data selection method brings additional performance gains, especially in scenarios with dual data heterogeneity.
\begin{figure}[t]
\centering
\includegraphics[width=0.9\columnwidth]{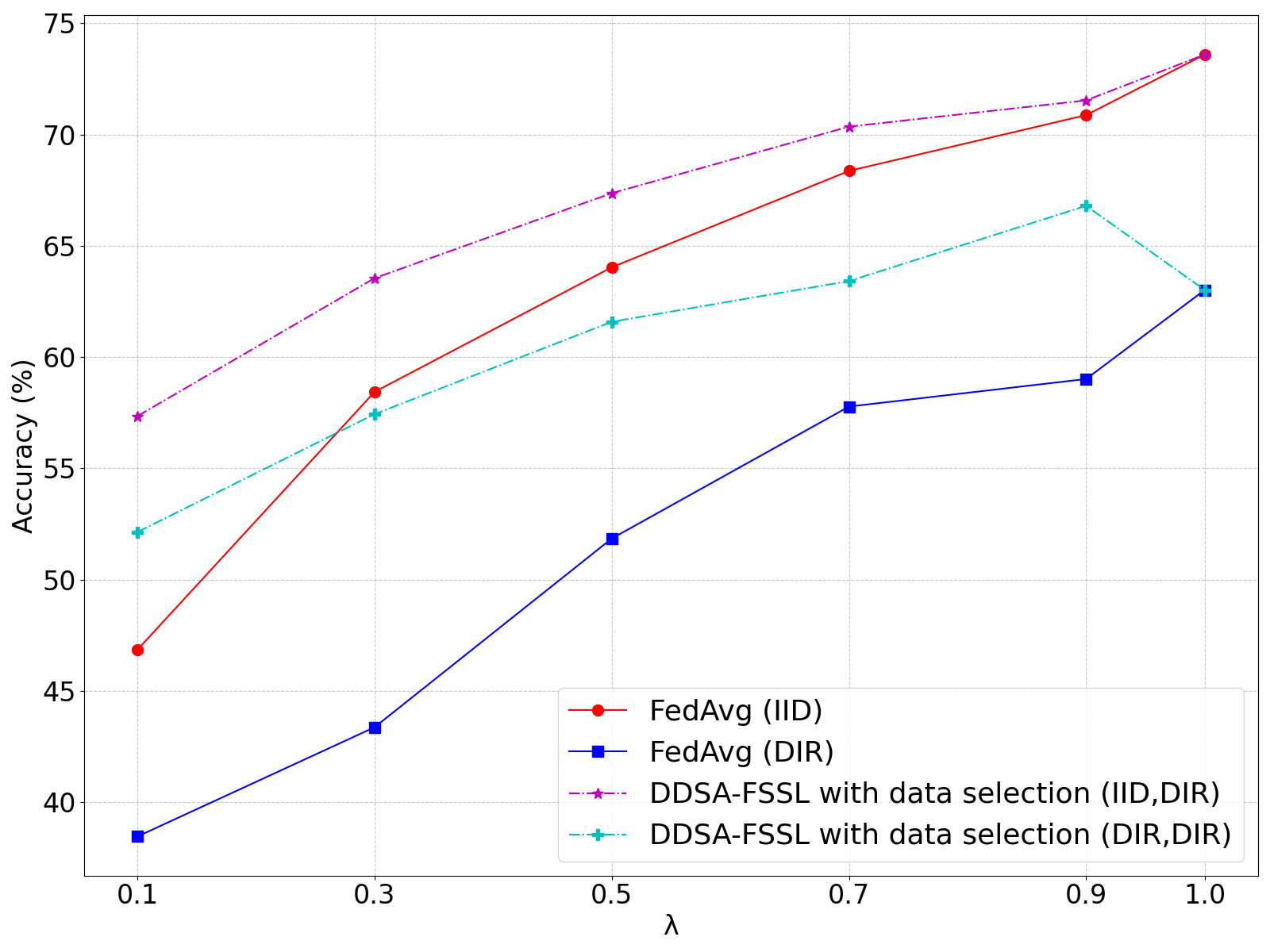}
\caption{The impacts of the ratio of labeled data on the performance under the condition of augmentation strength $\alpha=1$.}
\label{fig:change_lambda}
\vspace{-3mm}
\end{figure}

\begin{figure}[t]
\centering
\includegraphics[width=\columnwidth]{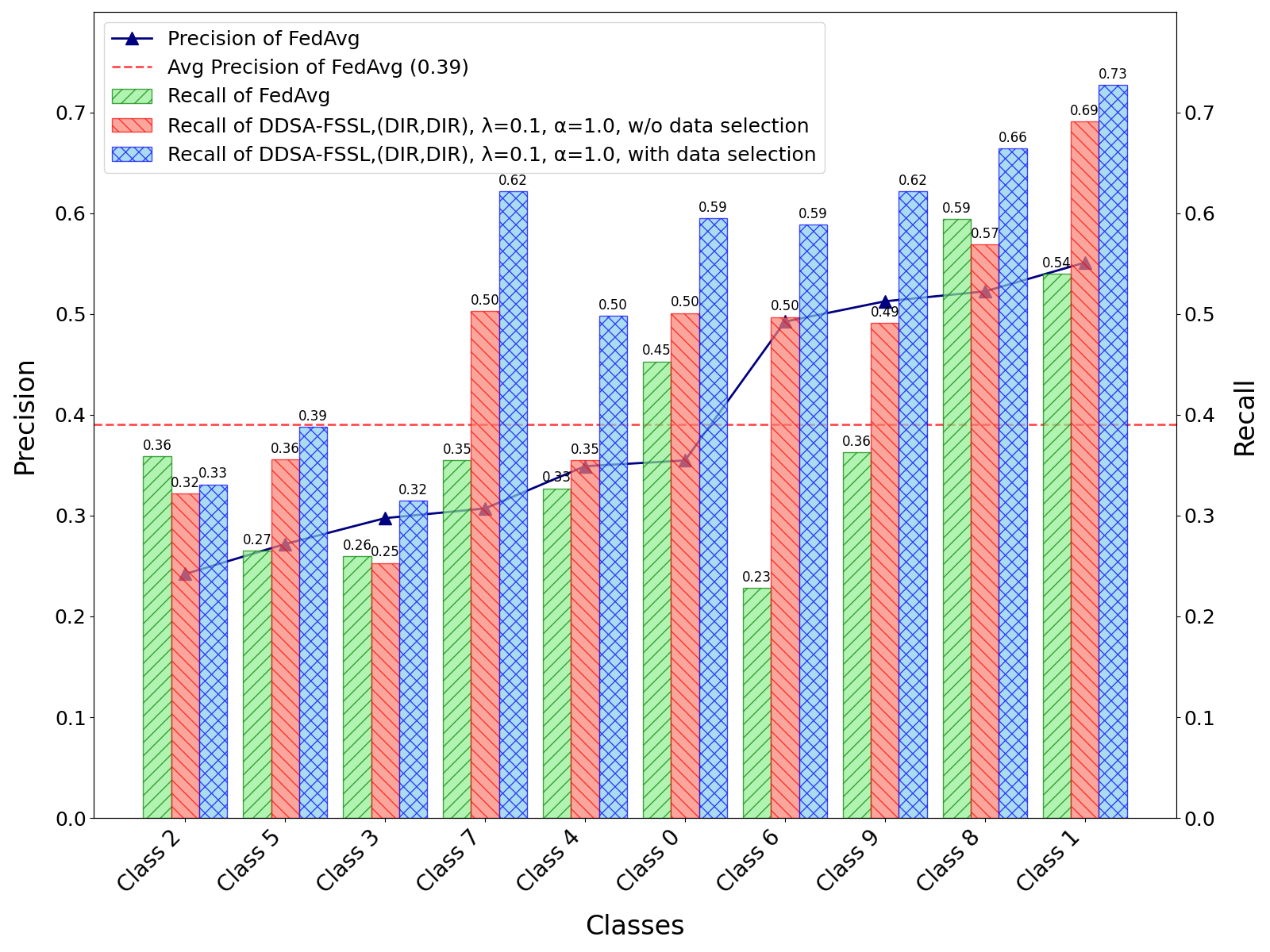}
\caption{Precision and recall variations across classes.}
\label{fig:dir_dir_matrix}
\end{figure}
Fig. \ref{fig:change_lambda} shows the performance of DDSA-FSSL with varying ratios $\lambda$ of labeled data under the condition of $\alpha=1$. When $\lambda=1.0$, all training data are labeled, which is equivalent to FedAvg-SL. The results indicate a positive correlation between $\lambda$ and classification accuracy, with DDSA-FSSL consistently outperforming the baseline. Notably, in scenarios with dual data heterogeneity, our method surpasses the FedAvg-SL results at $\lambda=0.7$ and $0.9$. This finding suggests that although synthetic data may not match the quality of real data, clients can effectively reduce data distribution heterogeneity by generating specific synthetic data, thereby improving performance.

Finally, we examine the changes in recall across different classes. In fact, the overall accuracy of the classifier on the dataset is equivalent to the average recall of each class. As shown in Fig. \ref{fig:dir_dir_matrix}, for CIFAR-10 dataset, the 10 classes are arranged in ascending order based on the precision obtained from the first step of DDSA-FSSL. It can be observed that after using the DDSA-FSSL method, classes with higher initial precision experience larger improvements in recall, while classes with lower precision show smaller or even negative changes in recall. This is because higher precision results in lower error rates during pseudo-labeling, leading to higher-quality data generated by c-LDM. Ablation studies further demonstrate that by specifically optimizing the label precision of data participating in c-LDM training, we achieved additional improvements in recall across all classes. Thus, reducing the error rate in pseudo-labeling is critical to the effectiveness of the proposed DDSA-FSSL.

\section{Conclusion}
In this paper, we propose a novel FSSL framework, DDSA-FSSL, based on DMs to tackle the challenges of data scarcity and non-IID distributions in FL. DDSA-FSSL utilizes collaboratively trained DMs to enable clients to generate synthetic data for missing classes of specific tasks. Additionally, ablation studies show that our proposed precision-optimized data selection method can improve the quality of the generated synthetic data, thereby leading to additional performance gains.

\footnotesize
\bibliographystyle{IEEEtran}
\bibliography{reference}{}

\end{document}